\documentclass{article}

\usepackage[preprint]{corl_2026}

\title{PATCH: Action-Chunk-Conditioned Latent Patch Innovation Monitoring for Robot Manipulation}

\author{Yanan Zhou \And Ranpeng Qiu \And Yincong Chen \And Jiajie Cui \And Weiming Zhi
\thanks{School of Computer Science, The University of Sydney.}
\thanks{Australian Centre For Robotics, The University of Sydney.}}

\usepackage{amsmath}
\usepackage{amssymb}
\usepackage{array}
\usepackage{booktabs}
\usepackage[font=small,labelfont=bf]{caption}
\usepackage{enumitem}
\usepackage{float}
\usepackage{graphicx}
\usepackage{wrapfig}
\usepackage{microtype}
\usepackage{cleveref}

\usepackage{cleveref}

\setlist[itemize]{leftmargin=1.15em,itemsep=0pt,topsep=1pt,parsep=0pt,partopsep=0pt}
\setlist[enumerate]{leftmargin=1.3em,itemsep=0pt,topsep=1pt,parsep=0pt,partopsep=0pt}
\AtBeginDocument{%
  \setlength{\abovedisplayskip}{4pt plus 1pt minus 2pt}%
  \setlength{\belowdisplayskip}{4pt plus 1pt minus 2pt}%
  \setlength{\abovedisplayshortskip}{2pt plus 1pt minus 1pt}%
  \setlength{\belowdisplayshortskip}{2pt plus 1pt minus 1pt}%
}

\newcommand{\method}{PATCH}
\newcommand{\router}{Router}

\begin{document}
\raggedbottom
\maketitle

\begin{abstract}
Learning-based manipulation policies have made substantial progress in real-world robot manipulation, particularly for short-horizon action generation. However, deployment in open workspaces remains fragile under unexpected local scene dynamics, such as moving objects, transient occlusions, or disturbances near the intended motion. Existing runtime monitors often rely on global observation anomalies, policy uncertainty, or frame-level visual changes, and struggle to distinguish task-relevant execution risk from benign visual variation. We introduce \method{}, an action-chunk-conditioned latent patch innovation monitor for deployment-time intervention. Given the active action chunk, \method{} defines a projected execution corridor, predicts latent patch evolution inside it, and accumulates persistent residuals unexplained by the robot's own motion. These residuals form a localized intervention signal that allows \method-\router{} to pause execution, select an available recovery source, and resume the original policy once localized innovation subsides. Experiments on real robot rollout data show that \method{} produces more stable and context-relevant triggers than competing runtime monitors. Real-robot deployment further demonstrates monitor-driven intervention and policy resumption for disturbance-aware manipulation. Project Page: \url{https://yananzhou5555.github.io/PATCH/}.
\end{abstract}

\section{Introduction}

Learning-based manipulation policies are moving from closed, single-skill settings toward more open-ended real-world autonomy. Methods such as Action Chunking with Transformers enable imitation policies to predict continuous action chunks for fine-grained manipulation~\cite{zhao2023learningfinegrainedbimanualmanipulation}. Recent robot policies further expand the range of objects, scenes, and task specifications that robots can handle~\cite{kim2024openvlaopensourcevisionlanguageactionmodel,intelligence2025pi05visionlanguageactionmodelopenworld}. However, open-workspace deployment remains sensitive to scene dynamics that training data cannot exhaustively cover, such as new objects entering the workspace, unexpected occlusions, and surrounding objects being moved. Runtime monitoring therefore needs to determine whether visual evolution is both abnormal for the active action chunk and localized to the region the policy will use, rather than merely detecting unusual observations.

\begin{figure}[t]
\centering
\vspace{-0.8em}
\includegraphics[width=1\linewidth]{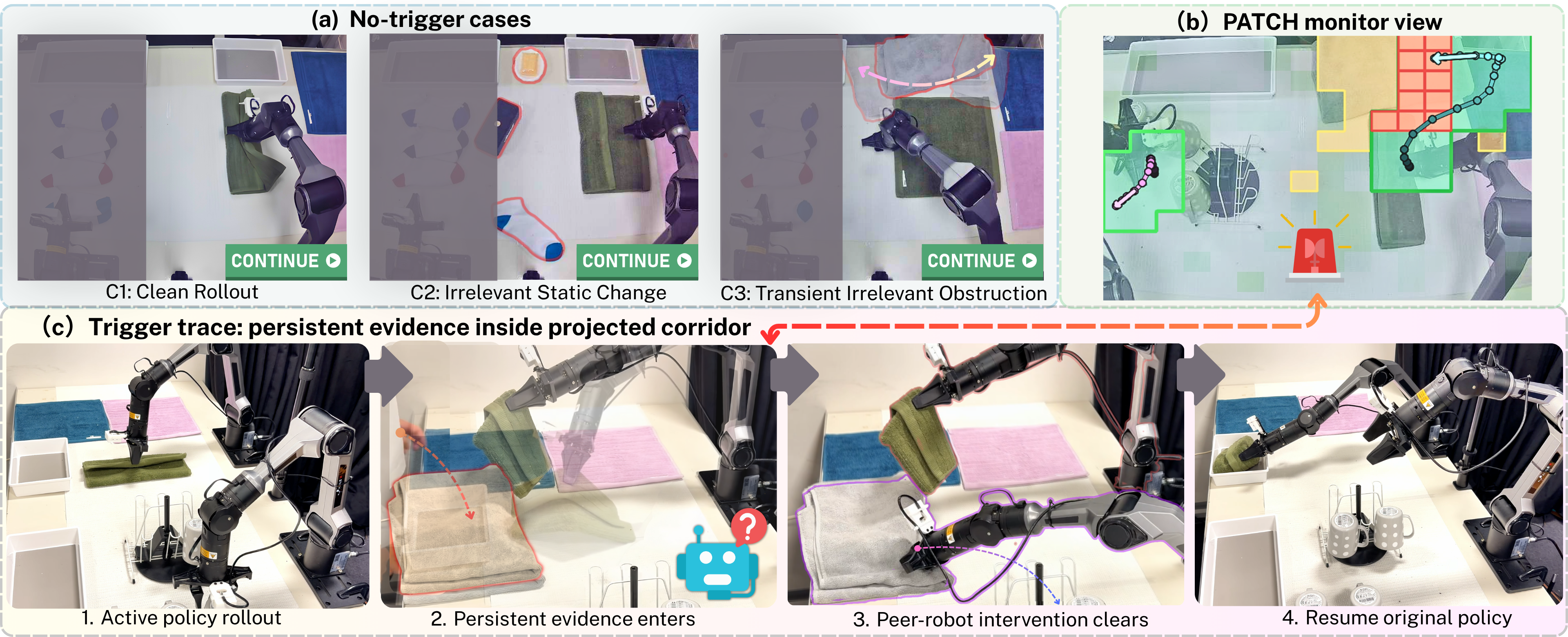}
\caption{
\textbf{\method{} intervention monitoring.}
(a) No-intervention cases that can confound visual-change monitors. (b) \method{} localizes action-relevant patch innovation inside the projected execution corridor. (c) Real rollout with \method-\router{} intervention and policy resumption.
}
\label{fig:teaser}
\vspace{-0.5em}
\end{figure}

Existing runtime monitors often rely on global observation anomalies, policy uncertainty, frame-level visual changes, or scalar failure scores. These signals can indicate distribution shift, low action confidence, broad visual change, or overall failure likelihood, but they do not reliably separate task-relevant execution risk from benign visual variation. \Cref{fig:teaser}(a) illustrates this limitation through cases that should not trigger intervention: C1 represents nominal execution; C2 introduces an irrelevant static change outside the action-relevant region; and C3 presents a transient obstruction without lasting execution impact. Execution-time monitoring therefore requires a stricter criterion: whether persistent external change lies inside the active policy's projected execution corridor and remains unexplained by normal action-conditioned evolution.

We address this with \method{}. Given the active action chunk, \method{} defines a projected execution corridor: the image or latent region traversed or affected by the policy's near-term motion. It predicts how latent patches inside this corridor should evolve under robot motion and treats the residual between predicted and observed evolution as patch innovation. Persistent innovation after self-motion filtering becomes the intervention signal. \method-\router{} then treats this localized signal as an interruption rather than a task failure: it preserves execution state, routes control to an available recovery source, and resumes the policy once localized innovation subsides.

The paper makes three contributions:
\begin{itemize}
    \item We formulate \emph{action-chunk-conditioned latent patch innovation monitoring}, which detects persistent external deviations inside the active policy's projected execution corridor rather than triggering on global observation change or policy uncertainty alone.

    \item We introduce \method{}, a latent patch monitor that predicts visual evolution inside this corridor, discounts self-motion-induced changes, and accumulates persistent unexplained residuals into deployment-time intervention signals.

    \item We evaluate \method{} on real robot rollout data with component ablations, and demonstrate \method-\router{} for monitor-driven intervention and policy resumption on real robots.
\end{itemize}

\section{Related Work}

\textbf{Action chunks and robot foundation policies.}
Imitation learning \cite{ravichandar2020recent, Diff_templates} enables complex motion patterns beyond motion plans \cite{Motion_planning, GeoFab_gloabL_opt} to specific goals. Short action sequences increasingly serve as the perception-control interface.
ACT and Diffusion Policy established chunked action prediction as a practical basis for high-bandwidth visuomotor manipulation~\cite{zhao2023learningfinegrainedbimanualmanipulation}.
Mobile ALOHA and real-time action-chunk execution scale this interface to mobile manipulation and low-latency control~\cite{fu2024mobilealohalearningbimanual}. TriPilot-FF \cite{li2026tripilot} advances this effort, enabling mobile manipulation data to be collected without being co-located, and torque information to be used in training. 
Open X-Embodiment, OpenVLA, and $\pi_{0.5}$ extend it across embodiments and tasks~\cite{open_x_embodiment_rt_x_2023,kim2024openvlaopensourcevisionlanguageactionmodel,intelligence2025pi05visionlanguageactionmodelopenworld}. Beyond action chunks, distributions over trajectory sequences \cite{zhi2023learning}, and stable dynamical systems have been used to represent action sequences \cite{periodic}.
In these works, the predicted sequence is primarily used to parameterize robot behavior.
\method{} uses the active chunk to construct a projected execution corridor, directing the monitor to visual regions the policy will traverse next and scene changes that may affect ongoing execution.

\textbf{Runtime monitoring, failure prediction, and latent dynamics.}
Runtime monitoring estimates reliability as observations change.
RND scores novelty through prediction error~\cite{burda2018explorationrandomnetworkdistillation}. Diffusion-based out-of-distribution signals have been studied in \cite{DBLP:journals/ral/ChengZMZJZ26}. Another line of work, ReDiffuser and robot-fleet world models use confidence or prediction error~\cite{pmlr-v235-he24e,liu2024multitaskinteractiverobotfleet}.
Failure-mode decomposition tracks progress and consistency~\cite{agia2024unpackingfailuremodesgenerative}.
FIPER and SAFE predict failures from action chunks or VLA features~\cite{romer2025failurepredictionruntimegenerative,gu2025safemultitaskfailuredetection}. Rewind-IL \cite{zheng2026rewind} introduces an action-chunk-based predictor that leverages conformal prediction and outperforms FIPER. Code-as-Monitor and RC-NF localize evidence with constraints or robot-conditioned normality~\cite{zhou2025codeasmonitorconstraintawarevisualprogramming,zhou2026rcnfrobotconditionednormalizingflow}.
FLARE, DreamGen, and DiWA use latent world models to frame prediction error as an execution counterfactual~\cite{corl2025flare,jang2025dreamgen,chandra2025diwa}.
\method{} uses latent dynamics to predict normal-execution patch states and accumulate persistent external innovation in the corridor.

\textbf{Monitor-driven intervention and assistance.}
SayCan and Inner Monologue ground help seeking in language and affordances~\cite{ahn2022icanisay,huang2022innermonologueembodiedreasoning}.
Robots Ask for Help studies assistance requests under uncertainty~\cite{ren2023robotsaskhelpuncertainty}.
Uncertainty Comes for Free links diffusion-policy uncertainty to human input, and UPS studies uncertainty-aware policy steering~\cite{he2025uncertaintycomesfree,yuan2026actasklearnuncertaintyaware}.
Dual-agent and latent theory-of-mind systems study peer-robot assistance~\cite{zhao2025dualagent,he2025latenttheoryminddecentralized}.
These approaches define the response after a trigger is produced.
Here, the trigger also carries localized evidence and active policy context.
\method-\router{} uses the intervention signal from \method{} with policy libraries and robot state to select an intervention mode and resume the original policy once localized evidence clears in the active region.

\begin{figure}[!t]
    \centering
    \vspace{-0.2em}
    \includegraphics[width=1\linewidth]{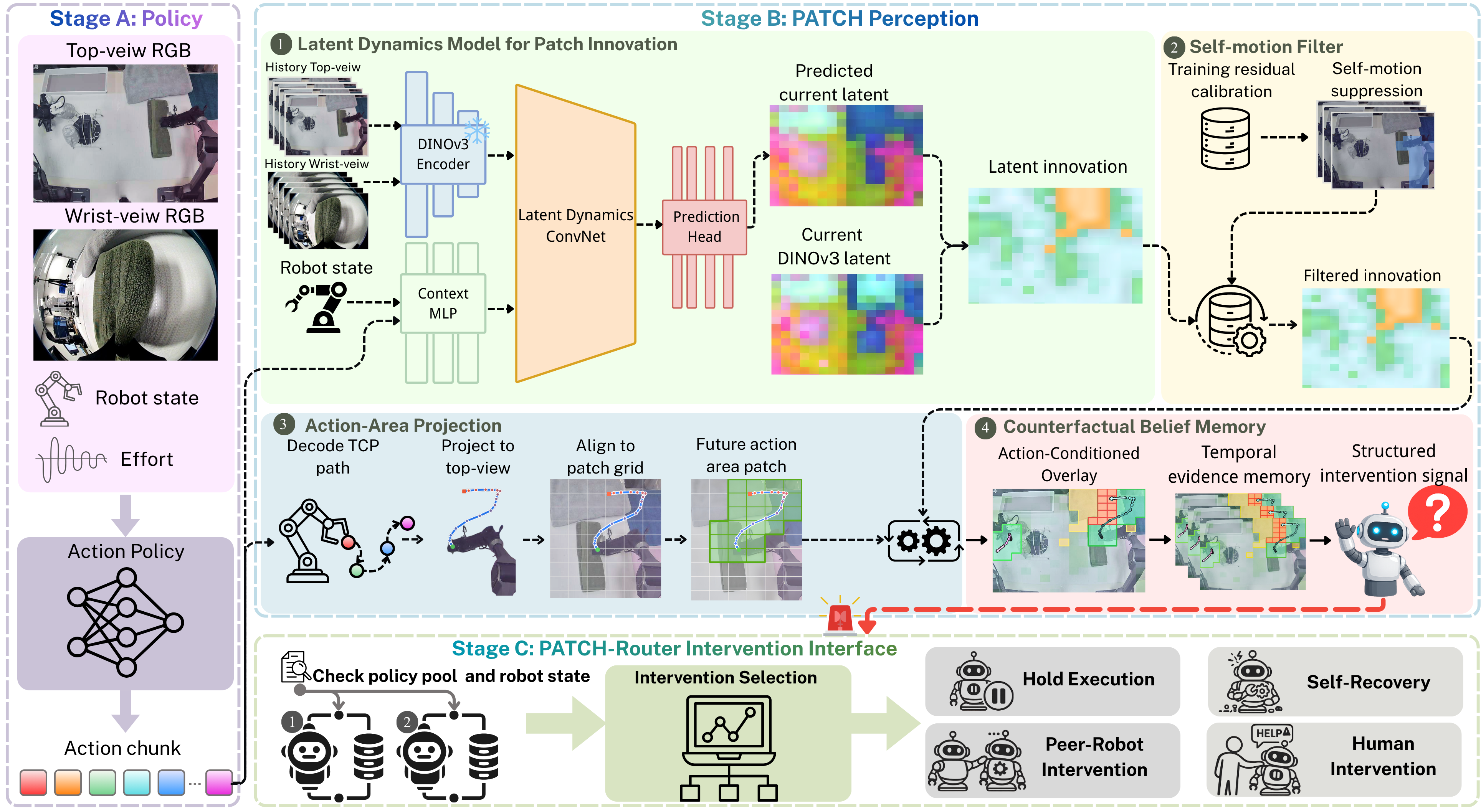}
    \caption{
    \textbf{The \method{} framework.}
    The monitor combines latent prediction, self-motion filtering, action-chunk projection, and evidence accumulation into \(\mathcal{I}_t\) for \method-\router{} hold and intervention selection.
    }
    \label{fig:pipeline}
    \vspace{-0.2em}
\end{figure}

\section{The PATCH Framework}

\subsection{Problem Formulation: Action-Chunk-Conditioned Latent Patch Innovation Monitoring}

We formulate \emph{action-chunk-conditioned latent patch innovation monitoring} as an execution-time problem for learned manipulation policies.
At each control step, the active policy identity \(\pi_t\) and action chunk \(U_t\) define the current policy context.
The action chunk induces a projected execution corridor: the image-space patch region expected to be traversed or affected by the robot over the short horizon.
\method{} should trigger only when this corridor contains persistent external latent innovation unexplained by normal robot-conditioned execution.
At time \(t\), the monitor receives
\begin{equation}
O_t =
\left(
I_t^{top},
I_t^{wrist},
q_t,
\tau_t,
\pi_t,
U_t
\right),
\qquad
U_t=\{a_t,\ldots,a_{t+H-1}\}.
\label{eq:input}
\end{equation}
Here, \(I_t^{top}\) and \(I_t^{wrist}\) are top-view and wrist-view RGB images, \(q_t\) is joint state, \(\tau_t\) is joint effort, \(\pi_t\) is the active policy identity, and \(U_t\) is its action chunk over horizon \(H\).
A monitored event must satisfy three conditions:
\emph{corridor relevance}, which restricts evidence to the corridor induced by \(U_t\);
\emph{external latent innovation}, which measures local latent deviation after accounting for robot self-motion; and
\emph{temporal persistence}, which suppresses transient changes through memory and latching.
\method{} maps \(O_t\) to a structured intervention signal containing a trigger, localized evidence region, evidence score, and active policy context.

\subsection{\method{} Monitor}

\method{} consists of four components: robot-conditioned latent dynamics, self-motion filtering, action-chunk corridor projection, and corridor evidence accumulation.

\vspace{-0.3em}
\textbf{Robot-Conditioned Latent Dynamics.}
This component predicts what the current top-view latent patch field should look like under normal execution, given recent belief history, wrist context, robot state, and the active action chunk.
The prediction defines the reference for patch-wise innovation: a patch is suspicious when its observed latent state is unlikely under normal dynamics for the active policy context.
We encode the top-view and wrist-view images with frozen dense visual encoders,
\(z_t=f_\phi(I_t^{top})\) and \(z_t^{wrist}=f_\phi^w(I_t^{wrist})\), where
\(z_t\in\mathbb{R}^{C\times h\times w}\).
The wrist stream provides robot-local context for expected motion and contact, while innovation is scored only in the top-view latent field.
Let \(b_t\) denote the monitor belief latent, updated in \Cref{eq:belief}.
A lightweight spatial ConvNet predicts a diagonal Gaussian over the current top-view latent:
\begin{equation}
(\mu_t,\sigma_t)
=
F_\theta
\left(
b_{t-K:t-1},
z^{wrist}_{t-K:t},
q_t,
U_t
\right),
\label{eq:prediction}
\end{equation}
with \(\sigma_t\) parameterized to be positive.
The model is trained on normal policy rollouts using synchronized visual, proprioceptive, and action sequences, without external-intervention labels, failure labels, or deployment-scene clean rollouts.
At test time, \method{} scores patch-wise latent innovation by the negative log-likelihood of the observed latent:
\begin{equation}
e_t(i,j)
=
\frac{1}{C}
\sum_{c=1}^{C}
-\log
\mathcal{N}
\left(
z_t(c,i,j);
\mu_t(c,i,j),
\sigma_t^2(c,i,j)
\right).
\label{eq:innovation}
\end{equation}
A high \(e_t(i,j)\) indicates that patch \((i,j)\) deviates from robot-conditioned normal latent evolution.

\vspace{-0.3em}
\textbf{Self-Motion Filtering.}
Self-motion filtering suppresses innovation caused by the robot's own execution, including arm motion, gripper motion, contact, held-object motion, and expected occlusion.
\method{} computes a self-motion support \(S_t\in[0,1]^{h\times w}\) by projecting robot geometry, gripper state, and the action-induced tool-center-point (TCP) sweep into the top-view image and aligning them with the latent patch grid.
If an object is held, its expected sweep is approximated by dilating the gripper-attached region along the projected TCP path.
Innovation is normalized using training residual statistics and filtered by \(S_t\):
\begin{equation}
\psi(e_t(i,j))
=
\operatorname{clip}
\left(
\frac{e_t(i,j)-\eta}{s_\eta},
0,
1
\right),
\qquad
X_t(i,j)
=
\psi(e_t(i,j))
\left(1-S_t(i,j)\right).
\label{eq:evidence}
\end{equation}
Here, \(\eta\) is a thresholded percentile of training residuals, and \(s_\eta\) is a validation-set residual scale used to map excess negative log-likelihood into \([0,1]\).
The output \(X_t\) remains high only where the observed latent is unlikely under normal dynamics and not explained by robot self-motion.

\vspace{-0.3em}
\textbf{Projected Execution Corridor from the Action Chunk.}
The projected execution corridor defines where external innovation is relevant to imminent execution.
\method{} decodes the active action chunk into an expected robot trajectory, maps it to a TCP path, projects it into the top-view image, dilates it for execution tolerance and end-effector extent, and aligns it with the latent patch grid:
\begin{equation}
\xi_t = FK(D_a(q_t,U_t)),
\qquad
P_t =
\mathcal{P}
\left(
\Pi_{top}(\xi_t)
\right)
\odot V,
\qquad
P_t\in[0,1]^{h\times w}.
\label{eq:corridor}
\end{equation}
Here, \(D_a\) maps the current joint state and policy actions to a predicted joint trajectory, \(FK(\cdot)\) computes the corresponding TCP path \(\xi_t\), \(\Pi_{top}\) projects this path into the top-view camera, \(\mathcal{P}(\cdot)\) performs dilation and patch-grid alignment, and \(V\) is an optional task or workspace region of interest.
The resulting \(P_t\) localizes action-relevant evidence to the corridor of the active chunk.

\vspace{-0.3em}
\textbf{Belief Memory.}
Belief memory converts filtered innovation into a stable prediction context.
It accepts observed latents where innovation is low or self-explained, and writes predicted normal latents where external innovation is active:
\begin{equation}
\alpha_t(i,j)
=
\max
\left(
1-\psi(e_t(i,j)),
S_t(i,j)
\right),
\quad
b_t(i,j)
=
\alpha_t(i,j)z_t(i,j)
+
\left(1-\alpha_t(i,j)\right)\mu_t(i,j).
\label{eq:belief}
\end{equation}
Thus, observed latents are retained when residuals are low or self-explained; otherwise the belief is filled with the predicted normal latent.
This prevents persistent external objects from becoming normal monitor context.

\vspace{-0.3em}
\textbf{Corridor Evidence and Latching.}
\method{} intersects external innovation evidence with the projected execution corridor:
\begin{equation}
R_t(i,j)=P_t(i,j)X_t(i,j),
\qquad
\rho_t =
\frac{
\langle R_t,P_t\rangle
}{
\langle P_t,P_t\rangle+\epsilon
}.
\label{eq:spatial}
\end{equation}
Here, \(R_t\) is patch-level evidence inside the projected corridor, and \(\rho_t\) is a corridor-normalized evidence score.
For binary \(P_t\), \(\rho_t\) is the mean external innovation inside the corridor; for soft \(P_t\), it is the corresponding corridor-weighted score.
Temporal accumulation and hysteresis produce the intervention trigger:
\begin{equation}
\bar{R}_t
=
\lambda \bar{R}_{t-1}
+
(1-\lambda)R_t,
\qquad
(y_t,\mathcal{K}_t)
=
\operatorname{Latch}_{\mathrm{int}}
\left(
\bar{R}_t,
\rho_t,
q_t,
\tau_t
\right).
\label{eq:certify}
\end{equation}
Here, \(y_t\) is the intervention trigger and \(\mathcal{K}_t\) is the localized evidence region.
The robot state and effort terms gate triggering and release during contact-rich execution, preventing release during high-effort contact and suppressing expected contact transients.
The latch rises after \(n_{\mathrm{on}}\) consecutive active frames, clears after \(n_{\mathrm{off}}\) release frames, and produces
\begin{equation}
\mathcal{I}_t =
\left(
y_t,
\mathcal{K}_t,
\rho_t,
\pi_t,
q_t,
\tau_t
\right).
\label{eq:intervention_signal}
\end{equation}
\(\mathcal{I}_t\) is the structured intervention signal passed to the downstream intervention interface.

\subsection{\method-\router{} Intervention Interface}

\method-\router{} consumes \(\mathcal{I}_t\) during real-robot deployment and operates at the intervention timescale rather than the control-loop timescale.
It reads the trigger \(y_t\), localized evidence region \(\mathcal{K}_t\), evidence score \(\rho_t\), and active policy context \((\pi_t,q_t,\tau_t)\).
Together, these fields specify whether intervention is needed, where it is needed, which policy is affected, and what robot state must be preserved during hold, intervention, and resumption.

Given self-recovery skills, peer-robot intervention libraries, and robot availability, the router selects
\begin{equation}
u_t^{router}
=
\operatorname{Select}
\left(
\mathcal{I}_t,
\mathcal{L}_t^{self},
\{\mathcal{L}_{t,m}^{peer}\}_{m=1}^{M},
\mathcal{A}_t^{robot}
\right),
\label{eq:router}
\end{equation}
where \(\mathcal{L}_t^{self}\) and \(\mathcal{L}_{t,m}^{peer}\) contain feasible intervention skills indexed by target region, and \(\mathcal{A}_t^{robot}\) encodes robot availability.
In our deployment, \(\operatorname{Select}(\cdot)\) is a priority-ordered feasibility test over
\[
u_t^{router}\in
\{\mathrm{hold},\mathrm{self\_recovery},\mathrm{peer\_robot\_int},\mathrm{human\_int}\}.
\]
If \(y_t\) is active and no recovery source has been selected, the acting robot enters \(\mathrm{hold}\) to preserve grasp, object state, and active policy context.
If a self-recovery skill can address \(\mathcal{K}_t\) without changing this preserved state, the router selects \(\mathrm{self\_recovery}\).
If the acting robot must remain in hold and another robot can reach \(\mathcal{K}_t\), it selects \(\mathrm{peer\_robot\_int}\).
If no robot-side recovery is feasible, it selects \(\mathrm{human\_int}\). The region \(\mathcal{K}_t\) provides the spatial target for intervention, while \(\rho_t\) and the latch state provide release conditions.
After \method{} observes that \(\mathcal{K}_t\) has cleared for the required release window, \method-\router{} releases hold and resumes the original policy.
Thus, resumption is triggered by local evidence clearance rather than by a fixed timeout or task reset.

\section{Empirical Evaluations}

\subsection{Experiment Setup and Evaluation Claims}

We evaluate \method{} on a real robot platform with two \emph{AgileX Piper} robot arms.
The manipulation policies are learned from real demonstrations using Action Chunking Transformer, and \method{} runs alongside the active policy during deployment.
At each step, \method{} receives the active action chunk and constructs the projected execution corridor used for intervention monitoring.
Detailed sensor streams, policy horizon, timestamp synchronization, and logging protocol are reported in \Cref{app:implementation}.

We evaluate three claims.
\textbf{Claim 1.} \method{} reduces false alarms relative to competing monitors while detecting action-relevant persistent disturbances.
\textbf{Claim 2.} Each main component of \method{} contributes to reliable trigger decisions.
\textbf{Claim 3.} \method{} produces intervention signals that support reliable \method-\router{} decisions when workspace changes affect real-robot execution.

\subsection{Action-Corridor Trigger Benchmark Setup}
\label{sec:trigger_eval}

\begin{figure}[t]
    \centering
    \captionsetup{skip=-0.3pt}
    \includegraphics[width=0.98\linewidth]{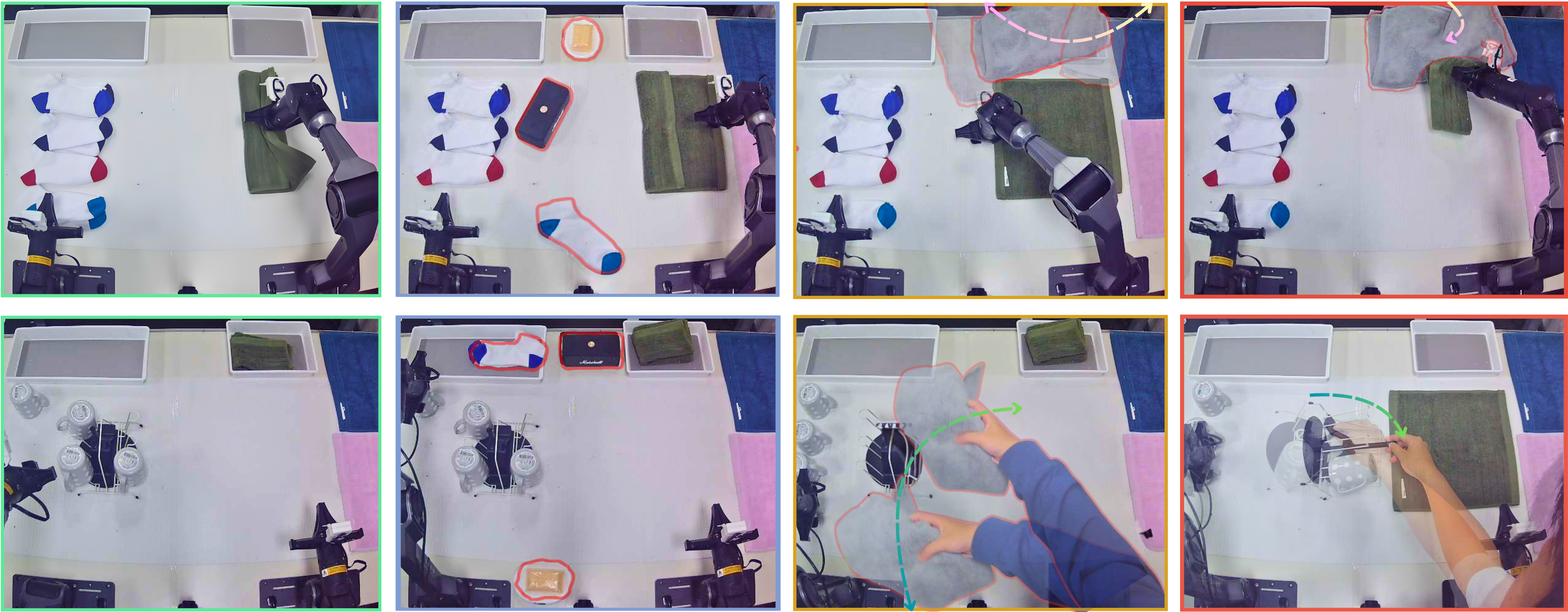}
    \vspace{0em}
    \begin{minipage}[t]{0.23\linewidth}
        \centerline{\footnotesize C1: Clean rollout}
    \end{minipage}\hfill
    \begin{minipage}[t]{0.23\linewidth}
        \centerline{\footnotesize C2: Static change}
    \end{minipage}\hfill
    \begin{minipage}[t]{0.23\linewidth}
        \centerline{\footnotesize C3: Transient obstruction}
    \end{minipage}\hfill
    \begin{minipage}[t]{0.23\linewidth}
        \centerline{\footnotesize C4: Persistent obstruction}
    \end{minipage}
    \vspace{0em}
    \caption{
    \textbf{Action-corridor trigger benchmark.}
    Top row: right-arm towel folding; bottom row: left-arm cup-to-rack.
    Columns show C1--C3 non-trigger cases and the C4 trigger case, testing nominal execution, irrelevant static change, transient obstruction, and persistent action-relevant obstruction.
    }
    \label{fig:data_classes}
\end{figure}

We evaluate Claim 1 with an offline action-corridor trigger benchmark, shown in \Cref{fig:data_classes}.
The benchmark contains two robot-task pairs, right-arm towel folding and left-arm cup-to-rack placement, evaluated under four event classes, denoted C1--C4.
Each evaluation window contains top-view RGB, the active robot's wrist-view RGB, robot state, joint effort, active policy identity, and active action chunk.
Each window is labeled according to whether the event should trigger intervention for the current policy execution state.

The four event classes are:
\emph{C1: clean execution}, where no external change occurs and any trigger is a false alarm;
\emph{C2: irrelevant static change}, where persistent clutter changes the observation but remains outside the projected execution corridor;
\emph{C3: transient obstruction}, where a towel or hand briefly enters the manipulation region and is then removed, producing short-lived visual change without lasting execution impact; and
\emph{C4: persistent action-relevant obstruction}, where an external obstacle remains inside the active policy's projected execution corridor.
C1--C3 are non-trigger classes, while C4 is the trigger class.
We collect 400 real-rollout evaluation windows balanced across the \(2\times4\) robot-task/event-class subsets.

%
%
%
\vspace{-0.3em}
\textbf{Baselines.}
We compare against four monitor families that isolate the main alternatives to \method{}: observation OOD, action confidence, scalar failure prediction, and dense patch change.
\textbf{PCA-kmeans}~\cite{liu2024multitaskinteractiverobotfleet} scores the distance from the current observation embedding to its nearest PCA-kmeans cluster.
\textbf{RND-A} adapts RND confidence estimation~\cite{pmlr-v235-he24e} and uses action-conditioned prediction error as an anomaly score.
\textbf{FIPER-style} follows runtime failure prediction in FIPER~\cite{romer2025failurepredictionruntimegenerative}, combining observation support and action-chunk reliability into a scalar trigger score.
\textbf{DINO-Latent Innovation} aggregates dense DINOv3 patch-level innovation~\cite{simeoni2025dinov3} into a frame-level trigger score.
All fitted models use the same \(D_{\mathrm{train}}\); scalar baselines additionally use \(D_{\mathrm{clean}}\) for clean-rollout correction.
\method{} trains its latent dynamics and residual evidence model from \(D_{\mathrm{train}}\), without additional labels.

\vspace{-0.3em}
\textbf{Metrics.}
Following monitor-level metrics used in FIPER~\cite{romer2025failurepredictionruntimegenerative}, we report false positive rate (FPR), true negative rate (TNR), true positive rate (TPR), and balanced accuracy (BalAcc) over 10 random seeds.
FPR is reported separately on C1--C3 to measure false alarms under clean execution, irrelevant static change, and transient obstruction.
TPR on C4 measures detection of persistent action-relevant obstructions.
BalAcc treats C4 as the positive class and C1--C3 as negative classes, unless otherwise specified.
Full metric definitions are provided in \Cref{app:metrics}.


\begin{table}[b]
\centering
\caption{
\textbf{Offline action-corridor trigger evaluation.}
Monitor baselines and \method{} ablations share the same event classes and metrics.
Values are mean $\pm$ std over 10 seeds; Clean calib. denotes clean-rollout calibration.
}
\label{tab:trigger_eval}
\scriptsize
\resizebox{\linewidth}{!}{%
\begin{tabular}{lccccccc}
\toprule
Method
& Clean calib.
& \(\mathrm{FPR}_{C_1}\downarrow\)
& \(\mathrm{FPR}_{C_2}\downarrow\)
& \(\mathrm{FPR}_{C_3}\downarrow\)
& \(\mathrm{TNR}\uparrow\)
& \(\mathrm{TPR}_{C_4}\uparrow\)
& \(\mathrm{BalAcc}\uparrow\) \\
\midrule
PCA-kmeans
& Yes
& $\mathbf{0.01 \pm 0.01}$
& $0.78 \pm 0.10$
& $0.90 \pm 0.07$
& $0.44 \pm 0.06$
& $\mathbf{0.99 \pm 0.01}$
& $0.72 \pm 0.04$ \\
RND-A
& Yes
& $\underline{0.02 \pm 0.01}$
& $\underline{0.08 \pm 0.05}$
& $\underline{0.20 \pm 0.09}$
& $\underline{0.90 \pm 0.04}$
& $0.32 \pm 0.18$
& $0.61 \pm 0.09$ \\
FIPER-style
& Yes
& $0.03 \pm 0.02$
& $0.26 \pm 0.10$
& $0.39 \pm 0.12$
& $0.77 \pm 0.07$
& $0.86 \pm 0.10$
& $\underline{0.82 \pm 0.06}$ \\
DINO-Latent Innovation
& No
& $0.09 \pm 0.04$
& $0.12 \pm 0.06$
& $0.97 \pm 0.03$
& $0.61 \pm 0.04$
& $\underline{0.98 \pm 0.01}$
& $0.80 \pm 0.03$ \\
\textbf{\method{} (ours)}
& No
& $0.05 \pm 0.03$
& $\mathbf{0.04 \pm 0.03}$
& $\mathbf{0.07 \pm 0.04}$
& $\mathbf{0.95 \pm 0.03}$
& $0.97 \pm 0.02$
& $\mathbf{0.96 \pm 0.02}$ \\
\midrule
DINO-Latent Innovation
& No
& $0.09 \pm 0.04$
& $0.12 \pm 0.06$
& $0.97 \pm 0.03$
& $0.61 \pm 0.04$
& $\mathbf{0.98 \pm 0.01}$
& $0.80 \pm 0.03$ \\
+ Projected Corridor
& No
& $0.05 \pm 0.03$
& $\mathbf{0.04 \pm 0.03}$
& $0.91 \pm 0.04$
& $0.68 \pm 0.04$
& $0.97 \pm 0.02$
& $0.83 \pm 0.03$ \\
ResNet-Latent Dynamics + Corridor
& No
& $0.06 \pm 0.02$
& $0.07 \pm 0.04$
& $0.42 \pm 0.09$
& $0.82 \pm 0.06$
& $0.84 \pm 0.08$
& $0.83 \pm 0.05$ \\
DINO-Latent Dynamics + Corridor
& No
& $0.05 \pm 0.03$
& $0.05 \pm 0.03$
& $0.31 \pm 0.08$
& $0.87 \pm 0.05$
& $0.94 \pm 0.04$
& $0.90 \pm 0.04$ \\
\textbf{\method{} (full)}
& No
& $0.05 \pm 0.03$
& $\mathbf{0.04 \pm 0.03}$
& $\mathbf{0.07 \pm 0.04}$
& $\mathbf{0.95 \pm 0.03}$
& $0.97 \pm 0.02$
& $\mathbf{0.96 \pm 0.02}$ \\
\bottomrule
\end{tabular}%
}
\end{table}

\vspace{-0.3em}
\textbf{Quantitative Results.}
\Cref{tab:trigger_eval} tests whether monitors distinguish visual novelty from action-corridor relevance.
PCA-kmeans clusters global observation embeddings and is therefore sensitive to scene-level mismatch between the deployment view and the fitted data distribution.
Its high \(\mathrm{TPR}_{C_4}\) comes with high \(\mathrm{FPR}_{C_2}\) and \(\mathrm{FPR}_{C_3}\), indicating that global visual change drives its triggers.
RND-A shows the opposite bias: it achieves high \(\mathrm{TNR}\), but its low \(\mathrm{TPR}_{C_4}\) suggests that action-confidence signals can remain stable even when an obstacle enters the projected execution corridor.
FIPER-style is the strongest scalar baseline because it combines observation support and action-chunk reliability, but it still depends on clean-rollout calibration to set its trigger threshold and retains elevated \(\mathrm{FPR}_{C_3}\) under transient dynamics.
DINO-Latent Innovation responds strongly to local visual dynamics, but its high \(\mathrm{FPR}_{C_3}\) shows that dense patch change without action-corridor restriction and temporal persistence over-triggers on transient obstructions.
\method{} preserves high \(\mathrm{TPR}_{C_4}\) while suppressing \(\mathrm{FPR}_{C_2}\) and \(\mathrm{FPR}_{C_3}\), supporting Claim 1: reliable intervention triggering requires evidence that is corridor-grounded, external, and temporally persistent.

%
%
%
%

\begin{figure}[t]
    \centering
    \vspace{-0.8em}
    \includegraphics[width=1\linewidth]{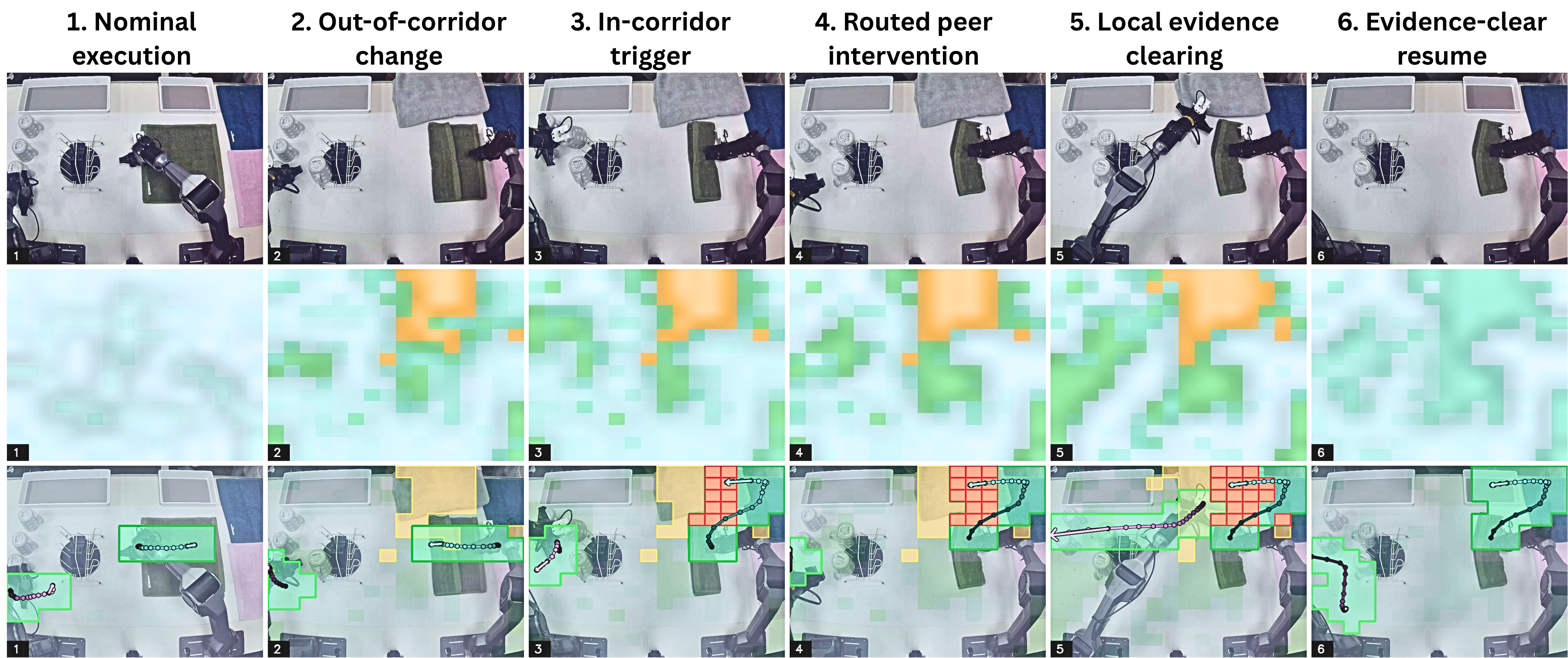}
    \caption{\textbf{\method{} evidence trace.} Top: rollout frames; middle: patch-level external innovation;
    bottom: projected execution corridors, action paths, and localized trigger evidence.}
    \label{fig:evidence_trace}
\end{figure}

\textbf{Component Ablations.} The lower block of \Cref{tab:trigger_eval} evaluates Claim 2 using the same event classes and metrics.
DINO-Latent Innovation maintains high \(\mathrm{TPR}_{C_4}\) but over-triggers on \(C_3\), showing that dense patch change alone is insufficient for selective intervention.
Adding the projected execution corridor reduces \(\mathrm{FPR}_{C_2}\) by restricting evidence to regions affected by the active action chunk.
Action-conditioned latent dynamics reduces \(\mathrm{FPR}_{C_3}\) by explaining normal robot-conditioned motion, while DINO latents retain stronger \(\mathrm{TPR}_{C_4}\) than ResNet latents under the same corridor constraint.
Full \method{} achieves the best balanced accuracy by combining corridor restriction, self-motion filtering, and temporal accumulation of external innovation inside the corridor. \Cref{fig:evidence_trace} visualizes these components over one synchronized rollout.
When external change appears outside the projected execution corridor, patch-level innovation rises but the corridor mask suppresses the trigger.
When an obstruction enters the corridor and persists, temporal accumulation maintains localized trigger evidence and \method{} issues the intervention signal.
After the peer robot clears the evidence region, accumulated evidence falls below the release threshold, and the original policy resumes under the preserved policy context.

\subsection{Robots that Assist}

\begin{figure}[t]
    \centering
    \includegraphics[width=1\linewidth]{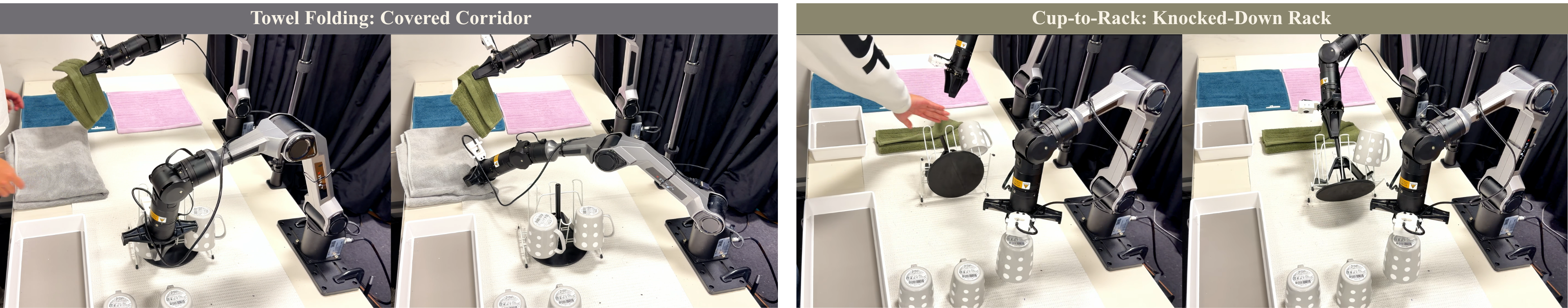}
    \vspace{0.05em}
    \begin{minipage}[t]{0.21\linewidth}
        \centerline{\footnotesize (a) Corridor obstruction}
    \end{minipage}\hfill
    \begin{minipage}[t]{0.21\linewidth}
        \centerline{\footnotesize (b) Peer clears obstruction}
    \end{minipage}\hfill
    \begin{minipage}[t]{0.21\linewidth}
        \centerline{\footnotesize (c) Rack knocked down}
    \end{minipage}\hfill
    \begin{minipage}[t]{0.21\linewidth}
        \centerline{\footnotesize (d) Peer restores rack}
    \end{minipage}
    \caption{\textbf{Real-world rollouts.} Two robots execute their active task policies in parallel; unexpected task-specific disturbances test \method{} monitoring and \method-\router{} closed-loop intervention decisions.}
    \vspace{-0.8em}
    \label{fig:real_world_rollouts}
\end{figure}

We evaluate Claim~3 on the real-world rollouts in \Cref{fig:real_world_rollouts}.
Two Piper arms execute independent policies in parallel: towel folding with the right arm and cup-to-rack placement with the left arm.
The rollouts introduce two task-specific disturbances: an obstacle entering the folding corridor, and a rack being knocked down so that the placement target becomes unavailable.
These cases test whether \method{} produces localized intervention evidence for the affected robot while suppressing irrelevant visual changes from the other active task.
They also test whether \method{}-\router{} can select an appropriate intervention mode, preserve the interrupted policy state, and resume execution once the localized evidence clears. Successful resumption after peer intervention indicates that the monitor remains tied to the active task context rather than frame-level visual change.


\begin{wraptable}[7]{l}{0.52\linewidth}
\centering
\vspace{-0.5em}
\captionsetup{width=\linewidth}
\caption{\textbf{Online real-robot stability.} All monitors share \method-\router{} and matched workspace changes.}
\label{tab:online_realrobot}
\scriptsize
\setlength{\tabcolsep}{1.2pt}
\renewcommand{\arraystretch}{1.04}
\begin{tabular*}{\linewidth}{@{\extracolsep{\fill}}lcccc@{}}
\toprule
Monitor & Inference & Intervention & Resume & Task \\
 & ms / Hz & Signal Acc. & Acc. & Completion Acc. \\
\midrule
FIPER-style & \textbf{3.94 / 254} & 80\% (16/20) & 35\% (7/20) & 30\% (6/20) \\
VLM monitor & 2850 / 0.35 & 85\% (17/20) & 20\% (4/20) & 15\% (3/20) \\
\textbf{\method{}} & 18.1 / 55.2 & \textbf{90\% (18/20)} & \textbf{90\% (18/20)} & \textbf{85\% (17/20)} \\
\bottomrule
\end{tabular*}
\end{wraptable}

We compare three monitors in closed-loop real-robot deployment under matched workspace disturbances. Since all methods use the same \method{}-\router{} interface, \Cref{tab:online_realrobot} isolates the monitor's effect on intervention timing, resumption, and task completion. Inference reports per-frame latency and throughput on a single RTX 5090 for local monitors, while the VLM row reports API round-trip latency; \method{} runs at the policy-matched 30 Hz control rate. Intervention Signal Acc. measures correct signaling for disturbances that block the active corridor, Resume Acc. measures correct clearance after obstruction removal, and Task Completion Acc. measures successful completion after resumption. FIPER-style detects many abnormal frames but struggles to identify clearance, reducing resume and completion accuracy. The VLM monitor often detects the blocking event, but its API latency can return after contact and required prompts tailored to expected accident categories. In contrast, \method{} updates localized corridor evidence at the policy rate, sustains intervention during persistent obstruction, verifies evidence-clear resumption, and achieves the highest task completion while preserving the active policy state.

\begin{wrapfigure}[11]{r}{0.485\linewidth}
    \vspace{-1em}
    \centering
    \captionsetup{width=\linewidth}
    \includegraphics[width=\linewidth]{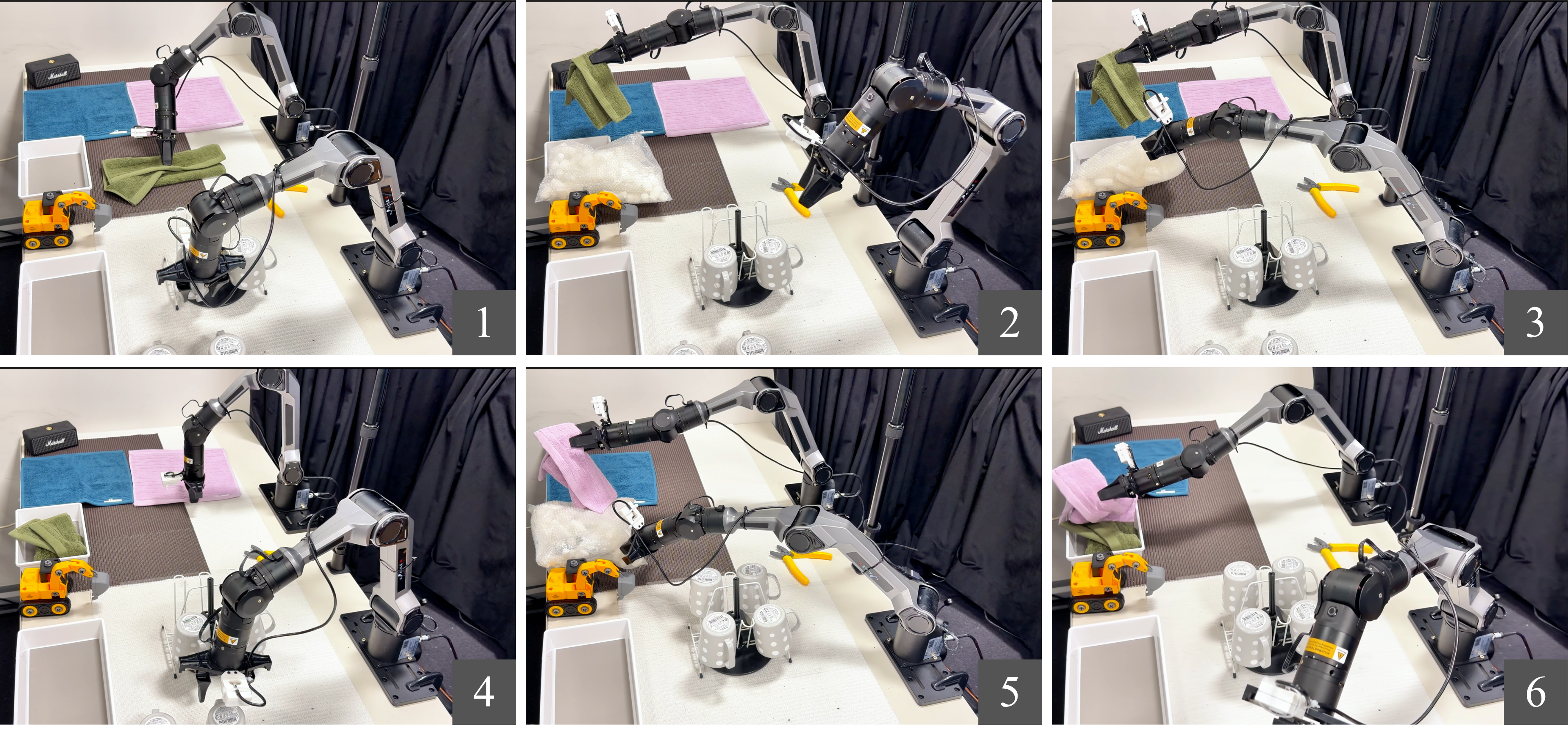}
    \vspace{-1em}
    \caption{\textbf{Unseen deployment.} New mat, clutter, and foam paper test action-corridor intervention.}
    \label{fig:ood_stress_rollout}
\end{wrapfigure}
\Cref{fig:ood_stress_rollout} further tests Claim~3 under unseen appearance and obstacle changes. The scene changes the mat, adds static clutter, and replaces the towel obstruction with foam paper. This separates appearance shift from intervention relevance: background changes and static objects remain continue cases because they do not intersect the active action corridor. Foam paper triggers intervention only after entering the projected corridor and persisting across the evidence window.

\section{Conclusion and Limitations}

We presented \method{}, an action-chunk-conditioned latent patch innovation monitor for deployment-time intervention.
\method{} uses active action chunks to form a short-horizon execution corridor, predict normal latent patch evolution, filter self-motion, and accumulate persistent external innovation for \method-\router{}.
Offline evaluation shows fewer false alarms on clean, irrelevant-change, and transient-obstruction cases while preserving recall on true execution blockers.
Ablations verify the roles of corridor projection, latent dynamics, self-motion filtering, and evidence memory.
Real-robot results show reliable intervention and policy resumption under workspace changes, including unseen backgrounds and obstacles.
Overall, action chunks provide useful perceptual structure for localized runtime monitoring and router decisions.

\noindent\textbf{Limitations.}
\method{} currently assumes a calibrated workspace observed from fixed camera viewpoints.
This makes the projected evidence region reliable in our tabletop setting, but also limits robustness to large viewpoint changes, camera motion, severe occlusion, and unstructured mobile operation where the monitored workspace may change over time. The projected execution corridor further depends on accurate action-to-TCP projection, robot-camera calibration, and workspace alignment. Extending \method{} to moving platforms or less structured scenes may require online calibration, e.g. with joint representation and calibration methods in \cite{JCR, tang2025bi}.

\bibliography{example}

\begin{thebibliography}{35}
\providecommand{\natexlab}[1]{#1}
\providecommand{\url}[1]{\texttt{#1}}
\expandafter\ifx\csname urlstyle\endcsname\relax
  \providecommand{\doi}[1]{doi: #1}\else
  \providecommand{\doi}{doi: \begingroup \urlstyle{rm}\Url}\fi

\bibitem[Zhao et~al.(2023)Zhao, Kumar, Levine, and Finn]{zhao2023learningfinegrainedbimanualmanipulation}
T.~Z. Zhao, V.~Kumar, S.~Levine, and C.~Finn.
\newblock {Learning Fine-Grained Bimanual Manipulation with Low-Cost Hardware}.
\newblock In \emph{Proceedings of Robotics: Science and Systems}, 2023.

\bibitem[Kim et~al.(2025)Kim, Pertsch, Karamcheti, Xiao, Balakrishna, Nair, Rafailov, Foster, Sanketi, Vuong, Kollar, Burchfiel, Tedrake, Sadigh, Levine, Liang, and Finn]{kim2024openvlaopensourcevisionlanguageactionmodel}
M.~J. Kim, K.~Pertsch, S.~Karamcheti, T.~Xiao, A.~Balakrishna, S.~Nair, R.~Rafailov, E.~P. Foster, P.~R. Sanketi, Q.~Vuong, T.~Kollar, B.~Burchfiel, R.~Tedrake, D.~Sadigh, S.~Levine, P.~Liang, and C.~Finn.
\newblock {OpenVLA: An Open-Source Vision-Language-Action Model}.
\newblock In \emph{Proceedings of The 8th Conference on Robot Learning}, pages 2679--2713, 2025.

\bibitem[Black et~al.(2025)Black, Brown, Darpinian, Dhabalia, Driess, Esmail, Equi, Finn, Fusai, Galliker, Ghosh, Groom, Hausman, Ichter, Jakubczak, Jones, Ke, LeBlanc, Levine, Li-Bell, Mothukuri, Nair, Pertsch, Ren, Shi, Smith, Springenberg, Stachowicz, Tanner, Vuong, Walke, Walling, Wang, Yu, and Zhilinsky]{intelligence2025pi05visionlanguageactionmodelopenworld}
K.~Black, N.~Brown, J.~Darpinian, K.~Dhabalia, D.~Driess, A.~Esmail, M.~R. Equi, C.~Finn, N.~Fusai, M.~Y. Galliker, D.~Ghosh, L.~Groom, K.~Hausman, B.~Ichter, S.~Jakubczak, T.~Jones, L.~Ke, D.~LeBlanc, S.~Levine, A.~Li-Bell, M.~Mothukuri, S.~Nair, K.~Pertsch, A.~Z. Ren, L.~X. Shi, L.~Smith, J.~T. Springenberg, K.~Stachowicz, J.~Tanner, Q.~Vuong, H.~Walke, A.~Walling, H.~Wang, L.~Yu, and U.~Zhilinsky.
\newblock {$\pi_{0.5}$: A Vision-Language-Action Model with Open-World Generalization}.
\newblock In \emph{Proceedings of The 9th Conference on Robot Learning}, pages 17--40, 2025.

\bibitem[Ravichandar et~al.(2020)Ravichandar, Polydoros, Chernova, and Billard]{ravichandar2020recent}
H.~Ravichandar, A.~S. Polydoros, S.~Chernova, and A.~Billard.
\newblock Recent advances in robot learning from demonstration.
\newblock \emph{Annual review of control, robotics, and autonomous systems}, 2020.

\bibitem[Zhi et~al.(2022)Zhi, Lai, Ott, and Ramos]{Diff_templates}
W.~Zhi, T.~Lai, L.~Ott, and F.~Ramos.
\newblock Diffeomorphic transforms for generalised imitation learning.
\newblock In \emph{Learning for Dynamics and Control Conference, {L4DC}}, 2022.

\bibitem[LaValle(2006)]{Motion_planning}
S.~M. LaValle.
\newblock \emph{Planning Algorithms}.
\newblock Cambridge University Press, USA, 2006.

\bibitem[Zhi et~al.(2023)Zhi, Akinola, van Wyk, Ratliff, and Ramos]{GeoFab_gloabL_opt}
W.~Zhi, I.~Akinola, K.~van Wyk, N.~Ratliff, and F.~Ramos.
\newblock Global and reactive motion generation with geometric fabric command sequences.
\newblock In \emph{IEEE International Conference on Robotics and Automation, ICRA}, 2023.

\bibitem[Fu et~al.(2025)Fu, Zhao, and Finn]{fu2024mobilealohalearningbimanual}
Z.~Fu, T.~Z. Zhao, and C.~Finn.
\newblock {Mobile ALOHA: Learning Bimanual Mobile Manipulation using Low-Cost Whole-Body Teleoperation}.
\newblock In \emph{Proceedings of The 8th Conference on Robot Learning}, pages 4066--4083, 2025.

\bibitem[Li et~al.(2026)Li, Zhou, Qiu, Wu, Ren, and Zhi]{li2026tripilot}
Z.~Li, Y.~Zhou, R.~Qiu, H.~Wu, G.~Ren, and W.~Zhi.
\newblock Tripilot-ff: Coordinated whole-body teleoperation with force feedback.
\newblock \emph{arXiv preprint arXiv:2602.09888}, 2026.

\bibitem[{Open X-Embodiment Collaboration}(2024)]{open_x_embodiment_rt_x_2023}
{Open X-Embodiment Collaboration}.
\newblock {Open X-Embodiment: Robotic Learning Datasets and RT-X Models}.
\newblock In \emph{2024 IEEE International Conference on Robotics and Automation (ICRA)}, pages 6892--6903, 2024.

\bibitem[Zhi et~al.(2024)Zhi, Zhang, and Johnson-Roberson]{zhi2023learning}
W.~Zhi, T.~Zhang, and M.~Johnson-Roberson.
\newblock Instructing robots by sketching: Learning from demonstration via probabilistic diagrammatic teaching.
\newblock In \emph{IEEE International Conference on Robotics and Automation}, 2024.

\bibitem[Zhi et~al.(2025)Zhi, Tang, Zhang, and Johnson-Roberson]{periodic}
W.~Zhi, H.~Tang, T.~Zhang, and M.~Johnson-Roberson.
\newblock Teaching periodic stable robot motion generation via sketch.
\newblock \emph{IEEE Robotics and Automation Letters}, 2025.

\bibitem[Burda et~al.(2019)Burda, Edwards, Storkey, and Klimov]{burda2018explorationrandomnetworkdistillation}
Y.~Burda, H.~Edwards, A.~Storkey, and O.~Klimov.
\newblock {Exploration by Random Network Distillation}.
\newblock In \emph{International Conference on Learning Representations}, 2019.

\bibitem[Cheng et~al.(2026)Cheng, Zheng, Ma, Zhang, Johnson{-}Roberson, and Zhi]{DBLP:journals/ral/ChengZMZJZ26}
H.~Cheng, T.~Zheng, Z.~Ma, T.~Zhang, M.~Johnson{-}Roberson, and W.~Zhi.
\newblock {DOSE3:} diffusion-based unified out-of-distribution detection on {\textdollar}{\textbackslash}mathbb\{SE\}(3){\textdollar} trajectories.
\newblock \emph{{IEEE} Robotics Autom. Lett.}, 11\penalty0 (2):\penalty0 1706--1713, 2026.
\newblock \doi{10.1109/LRA.2025.3640358}.
\newblock URL \url{https://doi.org/10.1109/LRA.2025.3640358}.

\bibitem[He et~al.(2024)He, Li, Li, Liu, and He]{pmlr-v235-he24e}
N.~He, S.~Li, Z.~Li, Y.~Liu, and Y.~He.
\newblock {ReDiffuser: Reliable Decision-Making Using a Diffuser with Confidence Estimation}.
\newblock In \emph{Proceedings of the 41st International Conference on Machine Learning}, pages 17921--17933, 2024.

\bibitem[Liu et~al.(2025)Liu, Zhang, Betala, Zhang, Liu, Ding, and Zhu]{liu2024multitaskinteractiverobotfleet}
H.~Liu, Y.~Zhang, V.~Betala, E.~Zhang, J.~Liu, C.~Ding, and Y.~Zhu.
\newblock {Multi-Task Interactive Robot Fleet Learning with Visual World Models}.
\newblock In \emph{Proceedings of The 8th Conference on Robot Learning}, pages 4286--4313, 2025.

\bibitem[Agia et~al.(2025)Agia, Sinha, Yang, Cao, Antonova, Pavone, and Bohg]{agia2024unpackingfailuremodesgenerative}
C.~Agia, R.~Sinha, J.~Yang, Z.~Cao, R.~Antonova, M.~Pavone, and J.~Bohg.
\newblock {Unpacking Failure Modes of Generative Policies: Runtime Monitoring of Consistency and Progress}.
\newblock In \emph{Proceedings of The 8th Conference on Robot Learning}, pages 689--723, 2025.

\bibitem[R{\"o}mer et~al.(2025)R{\"o}mer, Kobras, Worbis, and Schoellig]{romer2025failurepredictionruntimegenerative}
R.~R{\"o}mer, A.~Kobras, L.~Worbis, and A.~P. Schoellig.
\newblock {Failure Prediction at Runtime for Generative Robot Policies}.
\newblock In \emph{Advances in Neural Information Processing Systems 38}, 2025.

\bibitem[Gu et~al.(2025)Gu, Ju, Sun, Gilitschenski, Nishimura, Itkina, and Shkurti]{gu2025safemultitaskfailuredetection}
Q.~Gu, Y.~Ju, S.~Sun, I.~Gilitschenski, H.~Nishimura, M.~Itkina, and F.~Shkurti.
\newblock {SAFE: Multitask Failure Detection for Vision-Language-Action Models}.
\newblock In \emph{Advances in Neural Information Processing Systems 38}, 2025.

\bibitem[Zheng et~al.(2026)Zheng, Seenivasan, Johnson-Roberson, and Zhi]{zheng2026rewind}
G.~Zheng, S.~Seenivasan, M.~Johnson-Roberson, and W.~Zhi.
\newblock Rewind-il: Online failure detection and state respawning for imitation learning.
\newblock \emph{arXiv preprint arXiv:2604.16683}, 2026.

\bibitem[Zhou et~al.(2025)Zhou, Su, Chi, Zhang, Wang, Huang, Sheng, and Wang]{zhou2025codeasmonitorconstraintawarevisualprogramming}
E.~Zhou, Q.~Su, C.~Chi, Z.~Zhang, Z.~Wang, T.~Huang, L.~Sheng, and H.~Wang.
\newblock {Code-as-Monitor: Constraint-aware Visual Programming for Reactive and Proactive Robotic Failure Detection}.
\newblock In \emph{Proceedings of the IEEE/CVF Conference on Computer Vision and Pattern Recognition (CVPR)}, pages 6919--6929, 2025.

\bibitem[Zhou et~al.(2026)Zhou, Zhu, Yang, Zhao, Chen, and Jiang]{zhou2026rcnfrobotconditionednormalizingflow}
S.~Zhou, B.~Zhu, J.~Yang, X.~Zhao, J.~Chen, and Y.-G. Jiang.
\newblock {RC-NF: Robot-Conditioned Normalizing Flow for Real-Time Anomaly Detection in Robotic Manipulation}.
\newblock In \emph{Proceedings of the IEEE/CVF Conference on Computer Vision and Pattern Recognition (CVPR)}, 2026.

\bibitem[Zheng et~al.(2025)Zheng, Wang, Reed, Bjorck, Fang, Hu, Jang, Kundalia, Lin, Magne, Narayan, Tan, Wang, Wang, Xiang, Xu, Ye, Kautz, Huang, Zhu, and Fan]{corl2025flare}
R.~Zheng, J.~Wang, S.~Reed, J.~Bjorck, Y.~Fang, F.~Hu, J.~Jang, K.~Kundalia, Z.~Lin, L.~Magne, A.~Narayan, Y.~L. Tan, G.~Wang, Q.~Wang, J.~Xiang, Y.~Xu, S.~Ye, J.~Kautz, F.~Huang, Y.~Zhu, and L.~Fan.
\newblock {FLARE: Robot Learning with Implicit World Modeling}.
\newblock In \emph{Proceedings of The 9th Conference on Robot Learning}, pages 3952--3971, 2025.

\bibitem[Jang et~al.(2025)Jang, Ye, Lin, Xiang, Bjorck, Fang, Hu, Huang, Kundalia, Lin, Magne, Mandlekar, Narayan, Tan, Wang, Wang, Wang, Xu, Zeng, Zheng, Zheng, Liu, Zettlemoyer, Fox, Kautz, Reed, Zhu, and Fan]{jang2025dreamgen}
J.~Jang, S.~Ye, Z.~Lin, J.~Xiang, J.~Bjorck, Y.~Fang, F.~Hu, S.~Huang, K.~Kundalia, Y.-C. Lin, L.~Magne, A.~Mandlekar, A.~Narayan, Y.~L. Tan, G.~Wang, J.~Wang, Q.~Wang, Y.~Xu, X.~Zeng, K.~Zheng, R.~Zheng, M.-Y. Liu, L.~Zettlemoyer, D.~Fox, J.~Kautz, S.~Reed, Y.~Zhu, and L.~Fan.
\newblock {DreamGen: Unlocking Generalization in Robot Learning through Video World Models}.
\newblock In \emph{Proceedings of The 9th Conference on Robot Learning}, pages 5170--5194, 2025.

\bibitem[Chandra et~al.(2025)Chandra, Nematollahi, Huang, Welschehold, Burgard, and Valada]{chandra2025diwa}
A.~L. Chandra, I.~Nematollahi, C.~Huang, T.~Welschehold, W.~Burgard, and A.~Valada.
\newblock {DiWA: Diffusion Policy Adaptation with World Models}.
\newblock In \emph{Proceedings of The 9th Conference on Robot Learning}, pages 3378--3400, 2025.

\bibitem[Ichter et~al.(2023)Ichter, Brohan, Chebotar, Finn, Hausman, Herzog, Ho, Ibarz, Irpan, Jang, Julian, Kalashnikov, Levine, Lu, Parada, Rao, Sermanet, Toshev, Vanhoucke, Xia, Xiao, Xu, Yan, Brown, Ahn, Cortes, Sievers, Tan, Xu, Reyes, Rettinghouse, Quiambao, Pastor, Luu, Lee, Kuang, Jesmonth, Joshi, Jeffrey, Ruano, Hsu, Gopalakrishnan, David, Zeng, and Fu]{ahn2022icanisay}
B.~Ichter, A.~Brohan, Y.~Chebotar, C.~Finn, K.~Hausman, A.~Herzog, D.~Ho, J.~Ibarz, A.~Irpan, E.~Jang, R.~Julian, D.~Kalashnikov, S.~Levine, Y.~Lu, C.~Parada, K.~Rao, P.~Sermanet, A.~Toshev, V.~Vanhoucke, F.~Xia, T.~Xiao, P.~Xu, M.~Yan, N.~Brown, M.~Ahn, O.~Cortes, N.~Sievers, C.~Tan, S.~Xu, D.~Reyes, J.~Rettinghouse, J.~Quiambao, P.~Pastor, L.~Luu, K.-H. Lee, Y.~Kuang, S.~Jesmonth, N.~J. Joshi, K.~Jeffrey, R.~J. Ruano, J.~Hsu, K.~Gopalakrishnan, B.~David, A.~Zeng, and C.~K. Fu.
\newblock {Do As I Can, Not As I Say: Grounding Language in Robotic Affordances}.
\newblock In \emph{Proceedings of The 6th Conference on Robot Learning}, pages 287--318, 2023.

\bibitem[Huang et~al.(2023)Huang, Xia, Xiao, Chan, Liang, Florence, Zeng, Tompson, Mordatch, Chebotar, Sermanet, Jackson, Brown, Luu, Levine, Hausman, and Ichter]{huang2022innermonologueembodiedreasoning}
W.~Huang, F.~Xia, T.~Xiao, H.~Chan, J.~Liang, P.~Florence, A.~Zeng, J.~Tompson, I.~Mordatch, Y.~Chebotar, P.~Sermanet, T.~Jackson, N.~Brown, L.~Luu, S.~Levine, K.~Hausman, and B.~Ichter.
\newblock {Inner Monologue: Embodied Reasoning through Planning with Language Models}.
\newblock In \emph{Proceedings of The 6th Conference on Robot Learning}, pages 1769--1782, 2023.

\bibitem[Ren et~al.(2023)Ren, Dixit, Bodrova, Singh, Tu, Brown, Xu, Takayama, Xia, Varley, Xu, Sadigh, Zeng, and Majumdar]{ren2023robotsaskhelpuncertainty}
A.~Z. Ren, A.~Dixit, A.~Bodrova, S.~Singh, S.~Tu, N.~Brown, P.~Xu, L.~Takayama, F.~Xia, J.~Varley, Z.~Xu, D.~Sadigh, A.~Zeng, and A.~Majumdar.
\newblock {Robots That Ask For Help: Uncertainty Alignment for Large Language Model Planners}.
\newblock In \emph{Proceedings of The 7th Conference on Robot Learning}, pages 661--682, 2023.

\bibitem[He et~al.(2026)He, Cao, and Ciocarlie]{he2025uncertaintycomesfree}
Z.~He, Y.~Cao, and M.~Ciocarlie.
\newblock {Uncertainty Comes for Free: Human-in-the-Loop Policies with Diffusion Models}.
\newblock In \emph{2026 IEEE International Conference on Robotics and Automation (ICRA)}, 2026.

\bibitem[Yuan et~al.(2026)Yuan, Wu, and Bajcsy]{yuan2026actasklearnuncertaintyaware}
J.~Yuan, Y.~Wu, and A.~Bajcsy.
\newblock {When to Act, Ask, or Learn: Uncertainty-Aware Policy Steering}.
\newblock \emph{arXiv preprint arXiv:2602.22474}, 2026.

\bibitem[Zhao et~al.(2025)Zhao, Yue, Xie, Fang, Shao, and Guo]{zhao2025dualagent}
Z.~Zhao, X.~Yue, J.~Xie, C.~Fang, Z.~Shao, and S.~Guo.
\newblock {A Dual-Agent Collaboration Framework Based on LLMs for Nursing Robots to Perform Bimanual Coordination Tasks}.
\newblock \emph{IEEE Robotics and Automation Letters}, 10\penalty0 (3):\penalty0 2942--2949, 2025.

\bibitem[He et~al.(2025)He, Camps, Liu, Schwager, and Sartoretti]{he2025latenttheoryminddecentralized}
C.~He, G.~M.~S. Camps, X.~Liu, M.~Schwager, and G.~Sartoretti.
\newblock {Latent Theory of Mind: A Decentralized Diffusion Architecture for Cooperative Manipulation}.
\newblock In \emph{Proceedings of The 9th Conference on Robot Learning}, pages 392--405, 2025.

\bibitem[Sim{\'e}oni et~al.(2026)Sim{\'e}oni, Vo, Seitzer, Baldassarre, Oquab, Jose, Khalidov, Szafraniec, Yi, Ramamonjisoa, Massa, Haziza, Wehrstedt, Wang, Darcet, Moutakanni, Sentana, Roberts, Vedaldi, Tolan, Brandt, Couprie, Mairal, J{\'e}gou, Labatut, and Bojanowski]{simeoni2025dinov3}
O.~Sim{\'e}oni, H.~V. Vo, M.~Seitzer, F.~Baldassarre, M.~Oquab, C.~Jose, V.~Khalidov, M.~Szafraniec, S.~E. Yi, M.~Ramamonjisoa, F.~Massa, D.~Haziza, L.~Wehrstedt, J.~Wang, T.~Darcet, T.~Moutakanni, L.~Sentana, C.~Roberts, A.~Vedaldi, J.~Tolan, J.~Brandt, C.~Couprie, J.~Mairal, H.~J{\'e}gou, P.~Labatut, and P.~Bojanowski.
\newblock {DINOv3}.
\newblock \emph{Transactions on Machine Learning Research}, 2026.

\bibitem[Zhi et~al.(2024)Zhi, Tang, Zhang, and Johnson-Roberson]{JCR}
W.~Zhi, H.~Tang, T.~Zhang, and M.~Johnson-Roberson.
\newblock Unifying representation and calibration with 3d foundation models.
\newblock \emph{IEEE Robotics and Automation Letters}, 2024.

\bibitem[Tang et~al.(2025)Tang, Zhang, Johnson-Roberson, and Zhi]{tang2025bi}
H.~Tang, T.~Zhang, M.~Johnson-Roberson, and W.~Zhi.
\newblock Bi-manual joint camera calibration and scene representation.
\newblock \emph{arXiv preprint arXiv:2505.24819}, 2025.

\end{thebibliography}

\clearpage
\appendix

\section{Reproducibility Details}
\label{app:repro}

\subsection{Robot Platform and Sensing}
\label{app:platform}

Our experiments use two AgileX Piper robot arms in a shared tabletop workspace.
Each arm is equipped with a wrist-mounted fisheye RGB camera, and the workspace is observed by an external top-view RGB camera.
All cameras provide \(640\times480\) RGB streams at 30 Hz, matching the 30 Hz closed-loop policy execution rate.
For each rollout, the logging system records time-aligned top-view RGB, wrist-view RGB, robot state, joint effort, active policy identity, and the active action chunk.

\begin{table}[H]
\centering
\caption{\textbf{Robot platform and sensing setup.} The same sensing and logging interface is used for offline evaluation and closed-loop deployment.}
\label{tab:app_platform}
\scriptsize
\setlength{\tabcolsep}{4pt}
\begin{tabular}{p{0.28\linewidth}p{0.62\linewidth}}
\toprule
Item & Configuration \\
\midrule
Robot platform & Two AgileX Piper robot arms in a shared tabletop workspace \\
Wrist sensing & One wrist-mounted fisheye RGB camera per arm \\
External sensing & One top-view RGB camera observing the workspace \\
Camera streams & \(640\times480\) RGB at 30 Hz \\
Policy rate & 30 Hz closed-loop policy execution \\
Logged fields & Time-aligned top-view RGB, wrist-view RGB, robot state, joint effort, active policy id, active action chunk \\
\bottomrule
\end{tabular}
\end{table}

\subsection{Manipulation Policy Training}
\label{app:policy_training}

Both manipulation policies are trained from real robot demonstrations using the LeRobot ACT pipeline.
Each policy uses a ResNet18 visual backbone and a 7-layer decoder.
The action chunk size is 100, the action dimension is 7, and the robot state dimension is 14.
Top-view and wrist-view RGB images are resized to \(256\times256\) before being passed to the policy.

\begin{table}[H]
\centering
\caption{\textbf{ACT policy configuration.} Both base manipulation policies use the same LeRobot ACT architecture and input format.}
\label{tab:app_policy_config}
\scriptsize
\setlength{\tabcolsep}{4pt}
\begin{tabular}{p{0.28\linewidth}p{0.62\linewidth}}
\toprule
Item & Configuration \\
\midrule
Training pipeline & LeRobot ACT \\
Visual backbone & ResNet18 \\
Decoder & 7-layer decoder \\
Action chunk size & 100 \\
Action dimension & 7 \\
State dimension & 14 \\
Image input & \(256\times256\) top-view and wrist-view RGB images \\
Policy rate & 30 Hz \\
\bottomrule
\end{tabular}
\end{table}

\subsection{Base Policy Success Rate}
\label{app:base_policy}

Before evaluating intervention monitoring, we verify that the underlying ACT policies are reliable under undisturbed deployment.
Across 30 real-robot rollouts for each evaluated task, the base policies achieve over 90\% success without injected workspace disturbances.
This protocol keeps the closed-loop evaluation focused on monitor decisions under workspace disturbances and avoids attributing independent manipulation-policy failures to the monitor.

\subsection{Offline Benchmark Protocol}

\Cref{tab:app_offline_protocol} records the evaluation protocol behind \Cref{tab:trigger_eval}.
The purpose is to make the action-corridor benchmark auditable: the evaluated unit is an execution window with policy context, not an isolated image.

\begin{table}[H]
\centering
\caption{\textbf{Offline action-corridor protocol.} These controls prevent visual novelty, threshold fitting, and window sampling from leaking into the trigger decision.}
\label{tab:app_offline_protocol}
\scriptsize
\setlength{\tabcolsep}{4pt}
\begin{tabular}{p{0.25\linewidth}p{0.67\linewidth}}
\toprule
Item & Protocol \\
\midrule
Evaluation unit
& A window contains top-view RGB, wrist-view RGB, robot state, joint effort, active policy identity, and the action chunk \(U_t\). Labels attach to \((I_t^{top},I_t^{wrist},q_t,\tau_t,\pi_t,U_t)\). \\
Event ontology
& C1 clean, C2 persistent but outside the projected execution corridor, C3 transient inside the corridor, and C4 persistent inside the corridor. C1--C3 are continue classes; C4 is the trigger class. \\
Task balance
& The benchmark crosses two robot-task pairs with four event classes, producing \(2\times4\) robot-task/event subsets. We collect 400 windows balanced across these subsets. \\
Window sampling
& Fixed-duration windows are sampled from complete rollouts at the monitor rate. Windows with ambiguous repair or trigger-to-clear transitions are not used in the binary trigger table. \\
Training data
& \(D_{\mathrm{train}}\) fits representation models, latent dynamics, and PATCH residual calibration. External-intervention labels are not used for PATCH calibration. \\
Clean correction
& \(D_{\mathrm{clean}}\) is used only to set comparable operating points for scalar baselines that require clean-rollout threshold correction. \\
Randomization
& Reported table values average over ten random window-sampling seeds; the same sampled windows are used for all compared methods within a seed. \\
Leakage prevention
& Thresholds, residual percentiles, and router latches are fixed before evaluating benchmark windows. Evaluation labels are not used to adjust operating points after scoring. \\
\bottomrule
\end{tabular}
\end{table}

\subsection{Metric Definitions}
\label{app:metrics}

Following the monitor-level evaluation metrics used in FIPER~\cite{romer2025failurepredictionruntimegenerative}, the offline trigger benchmark reports false positive rate (FPR), true negative rate (TNR), true positive rate (TPR), and balanced accuracy (BalAcc).
For monitor predictions \(\hat y_t\in\{\mathrm{trigger},\mathrm{continue}\}\), we compute
\begin{align}
\mathrm{FPR}_{C_m}
&=
\Pr(\hat y_t=\mathrm{trigger}\mid C_m),
\quad m\in\{1,2,3\},\\
\mathrm{TNR}
&=
\Pr(\hat y_t=\mathrm{continue}\mid C_1\cup C_2\cup C_3),\\
\mathrm{TPR}_{C_4}
&=
\Pr(\hat y_t=\mathrm{trigger}\mid C_4),\\
\mathrm{BalAcc}
&=
\frac{1}{2}(\mathrm{TNR}+\mathrm{TPR}_{C_4}).
\end{align}
C1--C3 are continue classes, so \(\mathrm{FPR}_{C_m}\) measures false alarms under clean execution, irrelevant static change, and transient obstruction.
C4 is the trigger class, so \(\mathrm{TPR}_{C_4}\) measures detection of persistent execution-blocking obstruction.

\subsection{Baseline Implementation Details}
\label{app:baselines}

\Cref{tab:app_baselines} summarizes the implementation scope of monitors used in the offline benchmark or the online real-robot study.
Offline baselines use the same evaluated windows as \method{}.
Fitted representation or prediction models use \(D_{\mathrm{train}}\); methods marked with clean calibration in \Cref{tab:trigger_eval} use \(D_{\mathrm{clean}}\) only to set the operating point.

\begin{table}[H]
\centering
\caption{\textbf{Baseline implementation summary.} Each baseline isolates one common runtime-monitoring signal under matched evaluation inputs.}
\label{tab:app_baselines}
\scriptsize
\setlength{\tabcolsep}{4pt}
\begin{tabular}{p{0.20\linewidth}p{0.34\linewidth}p{0.36\linewidth}}
\toprule
Monitor & Fitting / calibration & Runtime score or decision \\
\midrule
PCA-kmeans
& Fits PCA observation embeddings and k-means clusters from \(D_{\mathrm{train}}\); uses \(D_{\mathrm{clean}}\) for operating-point calibration.
& Uses distance to the nearest cluster as an observation-level OOD score. \\
RND-A
& Trains a predictor to match a fixed random target network on action-conditioned inputs from \(D_{\mathrm{train}}\); uses \(D_{\mathrm{clean}}\) for calibration.
& Uses the resulting prediction error as the action-confidence anomaly score. \\
FIPER-style
& Follows the FIPER runtime-failure formulation using \(D_{\mathrm{train}}\) and clean-rollout calibration.
& Combines observation support and action-chunk reliability into one scalar trigger score. \\
DINO-Latent Innovation
& Uses the same frozen dense DINOv3 patch representation as the monitor ablation.
& Aggregates patch-level latent innovation into a frame-level dense-change score. \\
VLM monitor
& Uses no local fitting; prompts are fixed before evaluation.
& Queries a GPT-4o API with the current frame and policy context, then maps the response to the same router signal format for the online comparison. \\
\bottomrule
\end{tabular}
\end{table}

\subsection{\method{} Implementation Details}
\label{app:implementation}

\method{} uses a frozen DINOv3 ViT-B/16 encoder to extract dense visual tokens.
The \(16\times16\) patch grid gives 768-dimensional raw patch tokens, which are mapped by a fixed random projection and channel normalization to 128 latent channels.
The latent dynamics model is distinct from the ACT 7-layer policy decoder: it is a lightweight convolutional world model that predicts the current top-view latent patch state.
Given \(L=4\) previous top-view latents, \(L=4\) previous wrist-view latents, five robot states, and four recent action vectors, a context MLP maps the 98-dimensional proprioceptive-action history to 64 channels and broadcasts it over the \(16\times16\) grid.
The resulting 1088-channel tensor is processed by a \(1\times1\) convolution, two \(3\times3\) convolutions, and four residual convolution blocks with 256 hidden channels.
Two \(1\times1\) heads predict \(\mu_t\) and \(\log\sigma_t\), with delta prediction \(\mu_t=z_{t-1}^{top}+\Delta_\theta(\cdot)\).

\subsection{\method-\router{} Rule-Based Interface}
\label{app:router_rules}

\Cref{tab:app_router_rules} records the rule-based interface used in the real-robot demonstrations.
The router is a deployment component: it consumes \(\mathcal{I}_t\), checks recovery-library coverage and robot free state, and returns one of the intervention modes in \Cref{eq:router}.
The monitor remains responsible for trigger evidence and release evidence; the router only decides which free intervention source should act on the localized evidence region.
The key availability check is whether a robot is free to switch policy, which in our tasks mainly depends on whether its gripper is already holding an object; we mark a gripper as holding an object when its gripper torque exceeds 0.4.

\begin{table}[H]
\centering
\caption{\textbf{\method-\router{} rule-based interface.} The selector is intentionally simple: the acting robot holds its current state, and the router chooses a self or peer intervention source only if that robot is not already holding an object.}
\label{tab:app_router_rules}
\scriptsize
\setlength{\tabcolsep}{4pt}
\renewcommand{\arraystretch}{1.08}
\begin{tabular}{p{0.06\linewidth}p{0.86\linewidth}}
\toprule
Line & Rule \\
\midrule
1
& For a triggered robot \(i\), read \(\mathcal{I}_t^i=(y_t^i,\mathcal{K}_t^i,\rho_t^i,\pi_t^i,q_t^i,\tau_t^i)\) and let \(j\neq i\) denote the peer robot. \\
2
& If \(y_t^i=0\), keep executing the current task policy. If a previous intervention is latched and \(\mathcal{K}_t^i\) has cleared for the release window, release hold and resume the original policy. \\
3
& If \(y_t^i=1\), put robot \(i\) in \(\mathrm{hold}\) and preserve the current task state, active policy id, and cached action context. \\
4
& Compute \(\mathrm{free}(r)\) for each robot \(r\): the robot is not already executing an intervention policy, is not in a policy transition, and its gripper is empty, i.e., its gripper torque is at most 0.4 and it is not holding a task object. \\
5
& If \(\mathrm{free}(i)\) and a self-recovery skill can clear \(\mathcal{K}_t^i\), select \(\mathrm{self\_recovery}\). \\
6
& Else, if \(\mathrm{free}(j)\) and a peer intervention skill can clear \(\mathcal{K}_t^i\), keep robot \(i\) in \(\mathrm{hold}\) and select \(\mathrm{peer\_robot\_int}\). \\
7
& Else, if no robot-side intervention source is available, select \(\mathrm{human\_int}\) and expose the localized evidence target through the same interface. \\
8
& During a latched intervention event, keep the selected mode until \method{} reports that \(\mathcal{K}_t^i\) has cleared for the release window; then resume the original policy without reset. \\
\bottomrule
\end{tabular}
\end{table}

\subsection{Score and Threshold Audit}

\noindent\textbf{Audit checks.}
PATCH fixes the p99.5 residual threshold from \(D_{\mathrm{train}}\) before scoring C1--C4 windows and exports the realized \texttt{tau\_innov} field in per-frame logs.
Corridor overlap thresholds are fixed before reporting and audited through \texttt{overlap\_fraction}, \texttt{overlap\_count}, and \texttt{corridor\_count}.
Scalar baselines select thresholds using \(D_{\mathrm{clean}}\) only, then recompute class-wise metrics over the same ten window-sampling seeds as PATCH.
Every score CSV and replay overlay is stored with the source bag/session id; ROS message counts are not treated as trials.

\subsection{Additional Qualitative Rollouts}
\label{app:qualitative}

We provide additional qualitative demonstrations focused on \method-\router{} behavior.
These examples complement the real-world deployment results by showing how the same localized \method{} intervention signal can lead to different intervention modes depending on the acting robot state, peer free state, and recovery-library coverage.
Here, \(\mathrm{free}(r)\) follows the definition in \Cref{tab:app_router_rules}.

\begin{figure}[H]
    \centering
    \includegraphics[width=\linewidth]{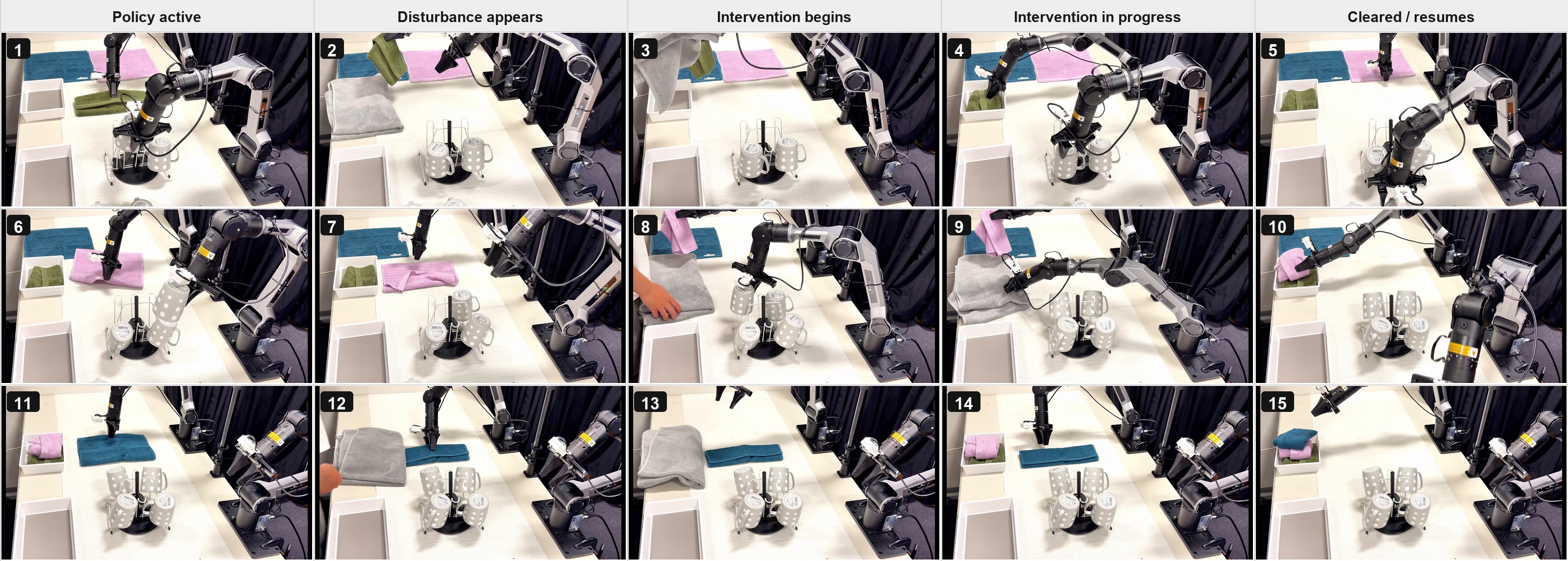}
    \caption{
    \textbf{Multi-intervention towel rollout.}
    Panels 1--15 are ordered frames from one continuous towel-folding deployment, grouped as three sequential intervention episodes.
    Episodes 1--5 and 6--10 use \(\mathrm{peer\_robot\_int}\) because robot \(i\) is occupied by the towel task \((\neg\mathrm{free}(i))\); episode 11--15 uses \(\mathrm{self\_recovery}\) because robot \(i\) is free.
    }
    \label{fig:app_towel_router_demo}
\end{figure}

\begin{table}[H]
\centering
\caption{
\textbf{\method-\router{} qualitative decisions.}
The demonstrations show how localized intervention signals map to different recovery modes under \(\mathrm{free}(\cdot)\) and policy-library constraints.
}
\label{tab:app_router_demo_summary}
\scriptsize
\setlength{\tabcolsep}{3pt}
\renewcommand{\arraystretch}{1.05}
\begin{tabular}{p{0.13\linewidth}p{0.25\linewidth}p{0.21\linewidth}p{0.18\linewidth}p{0.15\linewidth}}
\toprule
Episode & Disturbance & Acting robot state & Peer state & Router decision \\
\midrule
Towel 1--5 & Obstruction near folding region & \(\neg\mathrm{free}(i)\) & \(\mathrm{free}(j)\) & \(\mathrm{peer\_robot\_int}\) \\
Towel 6--10 & Obstruction near folding region & \(\neg\mathrm{free}(i)\) & \(\mathrm{free}(j)\) & \(\mathrm{peer\_robot\_int}\) \\
Towel 11--15 & Obstruction near folding region & \(\mathrm{free}(i)\) & Not queried & \(\mathrm{self\_recovery}\) \\
Cup-rack & Rack knocked down & \(\neg\mathrm{free}(i)\) & \(\mathrm{free}(j)\) & \(\mathrm{peer\_robot\_int}\) \\
\bottomrule
\end{tabular}
\end{table}

\begin{figure}[H]
    \centering
    \includegraphics[width=\linewidth]{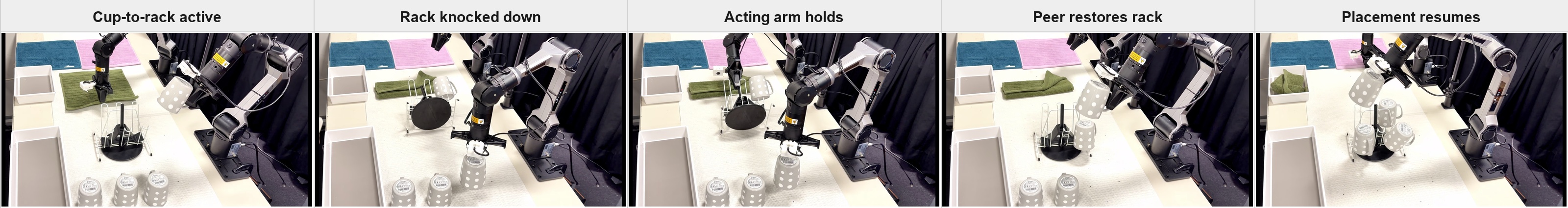}
    \caption{
    \textbf{Cup-rack peer intervention.}
    The cup-to-rack policy is active when the rack is knocked down.
    \method{} triggers on the affected placement target; the acting arm holds its execution state while \method-\router{} selects \(\mathrm{peer\_robot\_int}\), allowing the peer robot to restore the rack before policy resumption.
    }
    \label{fig:app_cup_rack_router_demo}
\end{figure}

If no robot-side recovery source is feasible, \method-\router{} selects \(\mathrm{human\_int}\).
This case occurs when no robot satisfies \(\mathrm{free}(r)\), or when the localized intervention signal does not match any self- or peer-recovery skill in the policy library.
We include this rule for completeness, while the qualitative demonstrations above focus on the robot-side peer-intervention and self-recovery modes used in our deployment.

\end{document}